\documentclass[10pt,twocolumn,letterpaper]{article}
\pdfoutput=1

\usepackage{iccv}
\usepackage{times}
\usepackage{epsfig}
\usepackage{graphicx}
\usepackage{amsmath}
\usepackage{amssymb}

\usepackage{booktabs}
\usepackage{adjustbox}
\usepackage{multirow}
\usepackage{comment}
\usepackage{authblk}
\usepackage[marginal]{footmisc}

\usepackage[breaklinks=true,bookmarks=false]{hyperref}

\usepackage[capitalize]{cleveref}
\crefname{section}{Sec.}{Secs.}
\Crefname{section}{Section}{Sections}
\Crefname{table}{Table}{Tables}
\crefname{table}{Tab.}{Tabs.}

\iccvfinalcopy 

\def\iccvPaperID{****} 

\ificcvfinal\fi

\begin{document}

\title{For A More Comprehensive Evaluation of 6DoF Object Pose Tracking}

\author[1]{Yang Li}
\author[1]{Fan Zhong}
\author[1]{Xin Wang}
\author[1]{Shuangbing Song}
\author[2]{Jiachen Li}
\author[1]{Xueying Qin}
\author[1]{Changhe Tu}
\affil[1]{School of Computer Science and Technology, Shandong University}
\affil[2]{College of Information Science and Electronic Engineering, Zhejiang University}
\renewcommand\Authands{ and }

\maketitle


\ificcvfinal\thispagestyle{empty}\fi

\begin{abstract}
Previous evaluations on 6DoF object pose tracking have presented obvious limitations along with the development of this area. In particular, the evaluation protocols are not unified for different methods, the widely-used YCBV dataset~\cite{xiangPoseCNN2018} contains significant annotation error, and the error metrics also may be biased. As a result, it is hard to fairly compare the methods, which has became a big obstacle for developing new algorithms.

In this paper we contribute a unified benchmark to address the above problems. For more accurate annotation of YCBV, we propose a multi-view multi-object global pose refinement method, which can jointly refine the poses of all objects and view cameras, resulting in sub-pixel sub-millimeter alignment errors. The limitations of previous scoring methods and error metrics are analyzed, based on which we introduce our improved evaluation methods. The unified benchmark takes both YCBV and BCOT~\cite{li2022bcot} as base datasets, which are shown to be complementary in scene categories. In experiments, we validate the precision and reliability of the proposed global pose refinement method with a realistic semi-synthesized dataset particularly for YCBV, and then present the benchmark results unifying learning\&non-learning and RGB\&RGBD methods, with some finds not discovered in previous studies.



\end{abstract}
\newcommand{\reffig}[1]{Figure~\ref{#1}}



\section{Introduction}
\label{sec:intro}

6DoF object pose tracking can obtain the accurate pose of 3D rigid objects, so it is important for many vision-based applications such as augmented reality, robotics, etc. Despite that many effective methods~\cite{stoiber2022iterative,dengPoseRBPF2021,wen2021bundletrack,wen2020se,zhongOcclusionAwareRegionBased3D2020,huangOcclusionAwareEdge2020,stoiber2021srt3d} have been proposed, so far it is still a challenge problem due to the high requirements for robustness and precision.

\begin{figure}
	\centering
	\setlength\tabcolsep{1pt}
	\small
	\begin{tabular}{cc}
		\includegraphics[width=0.49\linewidth]{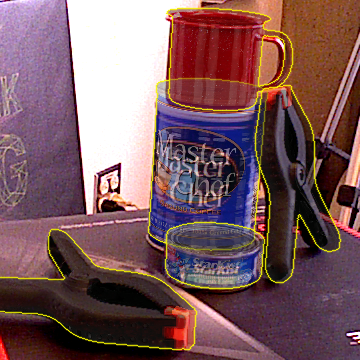} & \includegraphics[width=0.49\linewidth]{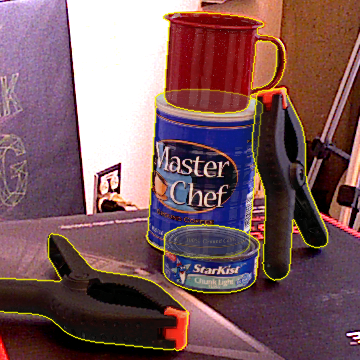} \\
		YCBV Original &  Our Refined
	\end{tabular}
	\caption{The original YCBV pose annotations (\emph{left}) and our refined results (\emph{right}). Object poses are visualized by blending the render image and highlighting the exterior contours. }
	\label{fig:intro}
\end{figure}

\begin{table*}
	\setlength\tabcolsep{4pt}
	\small
	\centering
	\begin{tabular}{ccccccccc}
		\toprule
		Method & Published &  Learning  & Main Dataset & Score & Error Metrics  & Init. Pose & Re-init. Condition & Re-init. Times  \\
		\midrule
		DeepIM~\cite{lideepim2018} & ECCV2018  & yes  & YCBV & AUC & ADD\&ADD-S & groundtruth & self-designed  & 290 \\
		PoseRBPF~\cite{dengPoseRBPF2021} & RSS2019  & yes  & YCBV & AUC & ADD\&ADD-S &  posecnn & self-designed & 2  \\
		TrackNet~\cite{wen2020se} & IROS2020   & yes & YCBV & AUC & ADD\&ADD-S & posecnn & self-designed & 2  \\
		ICG~\cite{stoiber2022iterative} & CVPR2022  & no  & YCBV & AUC & ADD\&ADD-S & groundtruth & not allowed & 0 \\
		RBOT~\cite{tjadenRegionBasedGaussNewtonApproach2019} & PAMI2018  & no & RBOT & SR & R-t error & groundtruth & error$>5cm5^\circ$ & - \\
		SRT3D~\cite{stoiber2021srt3d} & IJCV2022  & no & RBOT & SR & R-t error & groundtruth & error$>5cm5^\circ$ & - \\
		LDT~\cite{eccv2022} & ECCV2022 & no & RBOT & SR & R-t error & groundtruth & error$>5cm5^\circ$ & - \\
		\bottomrule
	\end{tabular}
	\caption{The evaluation datasets and methods of previous works. The re-init. times are reported from the original paper. }
	\label{tab:prev_evals}
\end{table*}

Since 6DoF pose tracking aims for high precision results and requires video inputs, it is particularly difficult to obtain high quality real-scene dataset for the evaluation~\cite{li2022bcot}. The main datasets used in previous methods are RBOT~\cite{tjadenRegionBasedGaussNewtonApproach2019} and YCBV~\cite{xiangPoseCNN2018}. 
RBOT is semi-synthesized with render objects floating on background image, and thus is unsuitable for the learning-based methods~\cite{lideepim2018,dengPoseRBPF2021,wen2020se} due to the domain generalization problem~\cite{tremblay2018training,rad2018feature}. YCBV is real-captured and commonly adopted by previous learning-based methods. However, as demonstrated in \cref{fig:intro} and noted in some previous works~\cite{haugaard2022surfemb,stoiber2022iterative,li2022bcot}, the pose annotation of YCBV is rather inaccurate because it is computed based on depth maps with significant depth error and color-depth misalignments, which makes it less reliable for high-precision tracking. 
The recently proposed BCOT dataset~\cite{li2022bcot} is captured in real scenes and precisely annotated. However, it contains only the case of single object, and meantime lacks the evaluations to previous learning-based methods, whose training processes are hard to be faithfully reproduced for new dataset. 


Besides, the evaluation protocols in previous works are not unified, as summarized in \cref{tab:prev_evals}. Learning-based methods are mainly evaluated on the YCBV dataset, with AUC score of ADD/ADD-S errors~\cite{hinterstoisser2012model} for ranking, while optimization-based methods are mainly evaluated on the RBOT dataset, with the success rate (SR) w.r.t the $5cm5^\circ$ error threshold for ranking. As a result, it is hard for the methods to be compared. Even within the same category, the ways for initialization, re-initialization, etc. are not all the same, which may result in unfair comparisons. 

The scoring methods and error metrics in previous works also present significant limitations for  comprehensive evaluation. For example, SR is known to be sensitive to the error threshold, and is unable to measure the continuous error distribution. We also find that the widely used ADD/ADD-S error is actually dominated by the translation error, and may overwhelm rotation error introducing obvious misalignments. As a result, biased scores and ranks may be produced, especially when both robustness and precision need to be considered.



In this paper we contribute methods and datasets to address the above problems. To improve the annotations of YCBV, we propose a multi-view multi-object global pose refinement method, namely the \emph{bundle pose refinement}, which can jointly optimize the object and camera poses in order for very small alignment errors.
We then introduce a new unified benchmark based on both YCBV and BCOT datasets, which are shown to be complementary in scene categories, and meantime suitable for both learning and non-learning methods. Limitations of previous scoring methods and error metrics are revealed and analyzed, based on which a new evaluation protocol is introduced. In experiments we first evaluate the proposed multi-view refinement method based on a realistic semi-synthesized YCBV variant with accurate groundtruth, and then present the benchmark results of  representative 6DoF pose tracking methods, with some discussions as well as validations to the proposed evaluation protocol. 



\section{Related Work}


\noindent\textbf{6DoF Object Pose Tracking} 
Recent 6DoF object pose tracking methods can be categorized as optimization-based or learning-based. The optimization-based methods~\cite{huangPixelWiseWeightedRegionBased2021,stoiber2020sparse,stoiber2021srt3d,zhongOcclusionAwareRegionBased3D2020} are studied mainly for augmented reality, which usually requires to run on mobiles, so computational efficiency is important and depth is usually not available. The learning-based methods~\cite{lideepim2018,wen2020se,dengPoseRBPF2021} are mainly studied in the area of robotics, for which powerful GPU and RGBD input can be utilized.

The main challenge of 6DoF pose tracking is to deal with textureless objects. The optimization-based methods usually take image edges~\cite{1017620,Harris1990RAPID,Vacchetti2004} or segmented region shapes~\cite{prisacariu_reid_bmvc2009,10.1007/s11263-015-0873-2,8237285} as the constraints. Early edge-based methods are known to be sensitive to background clutter~\cite{huangOcclusionAwareEdge2020,Seo2014}, which can be suspended by the region-based methods~\cite{prisacariu_reid_bmvc2009} thanks to the foreground segmentation. Recent progress of Stoiber et al.~\cite{stoiber2020sparse,stoiber2021srt3d, stoiber2022iterative} has showed that very fast speed and state-of-the-art accuracy can be achieved by a sparse region-based approach. The learning-based method is initially studied for 6DoF pose refinement~\cite{lideepim2018}, which is a problem closely related with 6DoF pose tracking. A specific learning-based tracking method then is introduced in se(3)-TrackNet~\cite{wen2020se}. In PoseRBPF~\cite{dengPoseRBPF2021}, the learned feature embedding is incorporated with particle filter~\cite{murphy2001rao} for global rotation estimation. This approach enables automatic initialization and re-initialization for the tracking, which are usually requires to be done with 6DoF pose estimation~\cite{xiangPoseCNN2018,pengPVNetPixelWiseVoting2019,su2022zebrapose}.

\noindent\textbf{Evaluation Datasets and Methods} \cref{tab:prev_evals} has summarized the main points regarding previous evaluations. Besides RBOT~\cite{tjadenRegionBasedGaussNewtonApproach2019} and YCBV~\cite{xiangPoseCNN2018}, there exist some other datasets, such as OPT~\cite{wu2017poster}, Choi~\cite{choi2013rgb}, YCBInEOAT~\cite{wen2020se}, which in comparison take smaller scales and less scene diversities, and thus are not commonly used. The accurate annotation of 6DoF object poses for videos is difficult in previous works, so markers are usually necessary~\cite{wu2017poster,hinterstoisser2012model,hodan2017t,liu2021stereobj}, which would contaminate the scene with unrealistic background. The calculation of pose error is not straightforward and could be application-dependent. Previous 6DoF tracking methods usually take R-t error~\cite{tjadenRegionBasedGaussNewtonApproach2019,li2022bcot} or ADD/ADD-S error~\cite{hinterstoisser2012model,xiangPoseCNN2018} as metrics. In \cite{hodavn2016evaluation}, metrics (i.e. VSD, ACPD, MCPD) specifically consider the effect of symmetric and occlusion for robotic grasping are proposed, which then is adopted in the BOP benchmark~\cite{hodan2018bop} for 6DoF pose estimation. We will further discuss related issues in \cref{sec:eval-method} and \cref{sec:benchmark}.

\noindent\textbf{Multi-view Pose Refinement} Most of previous 6DoF pose refinement methods~\cite{lideepim2018,iwase2021repose,lipson2022coupled,xu2022rnnpose} are for monocular image of single object, and are not accurate and reliable enough for pose annotation. 
%
In \cite{li2018unified} and \cite{li2022bcot}, multi-view constraint is exploited for solving single-view pose ambiguity, but they both require the camera poses to be known. In CosyPose~\cite{labbe2020cosypose}, the object and camera poses are jointly optimized. However, CosyPose just solves a consistent scene model and camera poses that can best explain its single-view pose estimations, which is fixed during the joint optimization, so errors of the single-view estimations could not be removed. As will be clarified in \cref{sec:bpr_init}, this is just the initialization of our method. The problem of joint object and camera poses estimation is also studied in object-level SLAM~\cite{salas2013slam++,yang2019cubeslam}, which however, are not specific for our problem and do not consider issues such as texutreless objects, high precision, etc. so the methods are also not ready for accurate pose annotation.




\section{Bundle Pose Refinement}
\label{sec:bundle}

This section will introduce the proposed global pose refinement method for high-precision pose annotation. We call this problem as \emph{bundle pose refinement}, akin the bundle adjustment for multi-view 3D reconstruction~\cite{triggs1999bundle}.

\renewcommand{\xi}{P}
\renewcommand{\nabla}{g}

\begin{figure}
	\centering
	\includegraphics[width=0.9\linewidth]{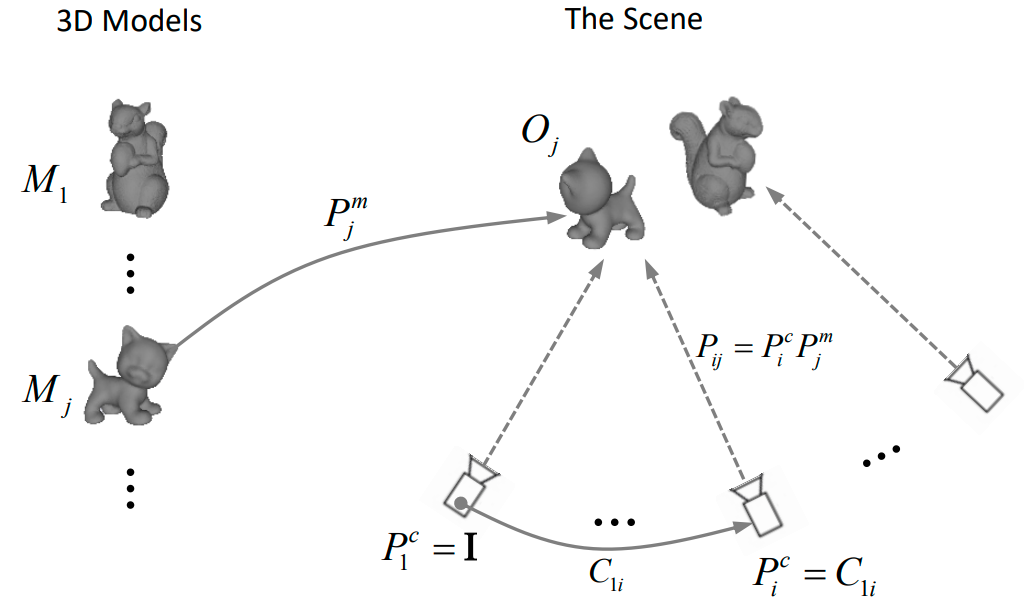}
	\caption{Bundle pose refinement. The scene model poses $\xi^m_j$ and camera poses $\xi^c_i$ are jointly optimized to improve the alignment with multi-view images.}
	\label{fig:bpr}
\end{figure}

\subsection{Inputs and Initialization} 
\label{sec:bpr_init}

Let $I_i, i=1,\cdots,n$ a sequence of images for the same scene, which contains $m$ different objects $O_j, j=1,\cdots,m$ with known 3D models $M_j$. The pose of $O_j$ in $I_i$ is $\xi_{ij}=\{R_{ij},t_{ij}\}$, which can be parameterized as a 6D vector~\cite{tjadenRegionBasedGaussNewtonApproach2019}. If $O_j$ is not appear in $I_i$, we denote it by $\xi_{ij}=\emptyset$. The inputs of our method include the image sequence $I_i$, the 3D models $M_j$, as well as the initial poses $\hat{\xi}_{ij}$, which can be obtained with single-view pose tracking.

We assume that the scene is static, or at least all objects can move as a whole. Therefore, $\xi_{ij}$ should meet the \emph{scene consistency} constraint, i.e. the relative pose between objects should keep the same for all views. Therefore, if $C_{ab}$ is the camera transformation from $I_a$ to $I_b$, then for all objects $j$ we should have $\xi_{bj}=C_{ab}\xi_{aj}$. As illustrated in \cref{fig:bpr}, in order to impose this constraint, we choose the camera framework of $I_1$ as the world coordinate system, then $\xi^m_j=\xi_{1j}$ and $\xi^c_i=C_{1i}$ should be the world object and camera poses respectively, and it is obvious that $\xi_1^c=\mathbf{I}$ and $\xi_{ij}=\xi^c_i\xi^m_j$.

Since the initial poses $\hat{\xi}_{ij}$ are usually not scene-consistent, in order to obtain the initial world poses $\hat{\xi}^m_j$ and $\hat{\xi}^c_i$, we can take the multi-view pose optimization method of CosyPose~\cite{labbe2020cosypose}, which minimizes the following cost:
\begin{equation}
	\hat{f}(\hat{\xi}^m_j, \hat{\xi}^c_i) = \sum_{i=1}^N \sum_{j,\hat{\xi}_{ij}\neq\emptyset}^M\sum_{X\in\mathcal{X}_j}\parallel \pi(\hat{\xi}_{ij}X)-\pi(\hat{\xi}^c_i\hat{\xi}^m_jX) \parallel
	\label{eq:init}
\end{equation}
where $\pi(.)$ is the camera projection function with known intrinsic parameters, $\mathcal{X}_j$ is a set of 3D points sampled from the surface of $M_j$. Minimizing $\hat{f}$ results in $\hat{\xi}^m_j, \hat{\xi}^c_i$ that can best explain the projections of $\hat{\xi}_{ij}$ for all views. 
Symmetric objects also can be handled by the method in \cite{labbe2020cosypose}.

Since \cref{eq:init} does not measure the alignments with images, it is obvious cannot remove the errors of $\hat{\xi}_{ij}$, for which purpose we need to further optimize $\xi^m_j$ and $\xi^c_i$. 

\subsection{Pose Refinement}
\label{sec:bpr_opt}

In order to achieve high-precision reliable pose refinement, a great challenge is to deal with textureless objects, for which there is no or very few stable 3D-2D matches can be found. On the other hand, since typical scenes are usually mixed with textured and textureless objects (e.g. YCBV), we need to fully make use of all available cues. Regarding this, a great limitation of state-of-the-art textureless tracking methods~\cite{stoiber2021srt3d,eccv2022} is that they make use of only object shape (exterior contours~\cite{eccv2022} or segmented regions~\cite{stoiber2021srt3d}) as cues, which  is obviously sub-optimal for many objects. Actually even for weak texture objects, interior regions can provide constraints that may be important for the accuracy. With these considerations, we propose to use the following cost function:
\begin{equation}
	\label{eq:totalcost}
	f(\Theta)=f^{edge}(\Theta)+\alpha f^{point}(\Theta)+\beta f^{t}(\Theta)
\end{equation}
in which $\Theta=\{\xi^c_i,\xi^m_j\}$ is the set of poses to be refined. $f^{edge}$ is an edge-based cost involving both exterior and interior edges. $f^{point}$ is a point matching cost that can provide more reliable matches for textured objects. $f^{t}$ is a temporal consistency cost measuring consistency between views. $\alpha$ and $\beta$ are constant weights for balancing the effects of different costs.

\begin{figure}
	\centering
	\setlength\tabcolsep{1pt}
	\small
	\begin{tabular}{ccc}
		\includegraphics[width=0.32\linewidth]{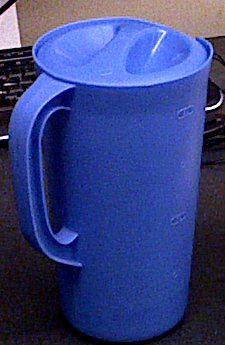} & 
		\includegraphics[width=0.32\linewidth]{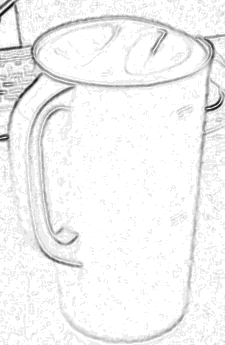} &
		\includegraphics[width=0.32\linewidth]{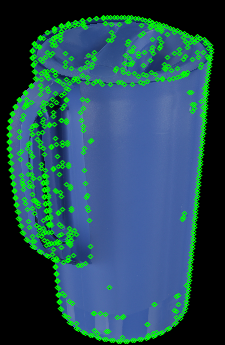} \\
		(a) $I_i$ clip for $O_j$ &  (b) $\bar{\nabla}_i$ clip & (c) $T_{ij}$ and $\mathcal{X}^e_{ij}$
	\end{tabular}
	\caption{The edge-based cost. Sampled model edges (c) are aligned with the continuous edge field (b) of the input image (a).}
	\label{fig:edgecost}
\end{figure}

In order to avoid the error-prone edge detection and matching processes as in previous edge-based methods~\cite{Seo2014,huangOcclusionAwareEdge2020}, we propose to use the following edge cost which takes continuous edge responses as loss
\begin{equation}
	f^{edge}(\Theta)=\sum_{i=1}^N \sum_{j,\xi_{ij}\neq\emptyset}^M\sum_{X\in\mathcal{X}^e_{ij}}\bar{\nabla}_i\left(\pi(\xi^c_i\xi^m_jX)\right) 
	\label{eq:Ee}
\end{equation} 
where $\bar{\nabla}_i$ is the edge map of $I_i$ as demonstrated in \cref{fig:edgecost}(b), which is computed as the  complement of the gradient magnitude normalized by the maximum value, so $\bar{\nabla}_i(p)$ is in the range $[0,1]$, and should be small for $p$ on strong edges.

As shown in \cref{fig:edgecost}(c), $\mathcal{X}^e_{ij}$ is a set of 3D points sampled from $T_{ij}$, i.e. the rendered template image of $O_j$ w.r.t the current pose $\xi_{ij}$. A maximum of 1000 points are sampled, with half of which for the exterior contours, and half for the interior edges from canny edge detection. For textureless object without obvious interior edges, the sampled points would be mainly on the exterior contours, which then would encode only the shape constraint, just as previous textureless tracking methods.

Note that for edges detected in $T_{ij}$ but not appear in $I_i$, a constant value about 1 would be taken for $\bar{\nabla}_i(p)$, whose effect then could be excluded without causing instability. This is different from previous methods requiring explicit edge matching~\cite{huangOcclusionAwareEdge2020}, for which mis-matching of edges may introduce large pose error. To overcome local minimum caused by the narrow band of edge response, coarse-to-fine search can be used. However, for the case as YCBV refinement, we find that it can work well even without the coarse-to-fine search.



The point matching loss is equal to solve a non-linear PnP problem~\cite{lepetit2009epnp} as
\begin{equation}
	f^{point} = \sum_{i=1}^N \sum_{j,\xi_{ij}\neq\emptyset}^M\sum_{X\in\mathcal{X}^p_{ij}}\parallel x-\pi(\xi^c_i\xi^m_jX) \parallel^\rho
	\label{eq:Ep}
\end{equation}
in which $X\in\mathcal{X}^p_{ij}$ is a 3D surface point, whose match in $I_i$ is $x$. To find the 3D-2D matches, we first find a set of 2D-2D matches $(x,x'), x\in I_i, x'\in T_{ij}$ by image matching (see below), then unproject $x'$ to the model space for the corresponding 3D point. $\rho<1$ is a parameter for robust estimation~\cite{eccv2022}, in our implementation it is fixed as 0.25.

The temporal term is added to ensure the consistency between views, so that matched pixels between different views can correspond to the same 3D model point
\begin{equation}
	f^{t}= \sum_{i=1}^N \sum_{k\in \mathcal{N}_i} \sum_{(x,x')\in\mathcal{M}_{ik}}\parallel x-\pi\left(\xi^c_i(\xi^c_k)^{-1}\pi^{-1}(x')\right) \parallel^\rho
\end{equation}
in which $\mathcal{N}_i$ is the set of neighbor views of $I_i$, $\mathcal{M}_{ik}$ is the set of matched points between $I_i$ and $I_k$, excluding those not on the objects. Note that the temporal consistency is independent with the scene model poses $\xi^m_j$, which thus are not involved in this term. $\rho$ is the same as that in \cref{eq:Ep}.

Both $f^{point}$ and $f^t$ require to find point matches between images. The matching method for them are the same. For a given image pair, we first calculate a dense match between them by the optical flow method of \cite{farneback2003two}, with reciprocal checking to suppress incorrect matches. We then detect corner points~\cite{shi1994good} on the source image, whose flow pairs are then taken as the resulting point matches. Compared with local feature matching~\cite{lowe2004distinctive}, this method can result in less outliers thanks to the smooth constraint of the optical flow. A maximum of 1000 matches are selected for both $f^{point}$ and $f^t$, but note that for textureless objects, only a few matches can be found because of the reciprocal checking and corner detection, so the effects of corresponding terms could be decreased automatically.

The total loss function $f(\Theta)$ is solved with Gauss-Newton method. To deal with long sequences such as that of YCBV, we first do global optimization in keyframes, and then interpolate remaining frames. Details about these issues can be found in the supplementary material.

\section{Evaluation Methods}
\label{sec:eval-method}


Besides the datasets, previous evaluation methods also present significant limitations. In the following we will analyze related issues, and then propose our improvements.

\subsection{The Scores}
\label{sec:scoringm}

Previous methods usually take SR (success rate)~\cite{tjadenRegionBasedGaussNewtonApproach2019,stoiber2021srt3d,eccv2022} or AUC (area under curve)~\cite{wu2017poster,lideepim2018,stoiber2022iterative} for scoring the accuracy of resulting poses. 

The SR score is proportional to the number of successfully tracked frames. In the evaluation protocol of RBOT and BCOT benchmarks, after each tracking loss the pose would be re-initialized by the groundtruth. Therefore, the  re-initialization times is proportional to 1-SR. The tracking loss needs to be determined by an error threshold (e.g. $5cm5^\circ$). Since all errors below the threshold are taken as the same, SR could not measure the precision of tracking methods, and meantime is sensitive to the error threshold. In the BCOT benchmark, different error thresholds (e.g. $2cm2^\circ$, $0.05d$-ADD) are tested, which however, would make the ranking complicated. 



The AUC score as introduced in ~\cite{wu2017poster} can measure the continuous error distribution of the resulting poses. Unfortunately, in contrast with SR, the re-initialization times is not taken into account by AUC. Since 6DoF tracking methods usually take groundtruth or 6D pose estimation results for the re-initialization, it is obvious that the methods with more times of re-initializations (lower SR) would be beneficial. This is obviously unreasonable. As shown in \cref{tab:prev_evals}, previous methods for the re-initialization are quite different, which makes the resulting AUC scores less reliable to be compared. 

To address the above problems, a simple way is to use AUC while prohibiting re-initialization. However, for datasets with only a few long test sequences (e.g. RBOT and YCBV), this approach may overemphasize some occasional cases (e.g. heavy occlusion) that would cause tracking losses. Therefore, we propose to pre-segment the test videos as a set of subsequences with different length, and then test using subsequences without re-initialization. In this way the datasets can be better make used. More details will be introduced in \cref{sec:subseq}.

\subsection{The Error Metrics}
\label{sec:errorm}

In previous works the most commonly used error metrics are the R-t error~\cite{tjadenRegionBasedGaussNewtonApproach2019} and the ADD(or ADD-S for symmetric objects)  error~\cite{hinterstoisser2012model}. The R-t error is intuitive, but because the rotation and translation errors are separated, it is inconvenient for computing the AUC score. ADD eliminates this problem by measuring the deviations of transformed model points, and thus is preferred for calculating AUC score.

Nevertheless, since for monocular tracking, the translation errors in the $Z$ direction ($tz$) are usually much larger than that in the $X$ and $Y$ directions ($tx$ and $ty$), it would be somewhat difficult to determine the AUC error bound. For instance, the error bound of ADD is usually taken as $10cm$ for YCBV, which is obviously too large for $tx$ and $ty$ (corresponds to reprojection errors over 100 pixels). On the other hand, for $tz$ it is not too large in order to measure the tracking robustness. As a result, the ADD-AUC scores reported in previous works are actually dominated by the translation errors, and are insensitive to rotation errors that may have introduced obvious mis-alignments.


Therefore, we propose to use reprojection (PRJ) error at the same time. According to our above analysis, PRJ is a nice compensation of ADD, and meantime ADD also can compensate the insensitivity of PRJ to the errors of $tz$. Considering application requirements, robotic grasping should be more concern about ADD error, while AR should be more concern about PRJ error. Using them both thus can result in more comprehensive scores and ranks.

\section{The Unified Benchmark}
\label{sec:benchmark}

Based on the above methods and analysis, we introduce a unified benchmark in order for a comprehensive evaluation of 6DoF tracking.

 \begin{table}
	\setlength\tabcolsep{2pt}
	\small
	\centering
	\begin{tabular}{lcccccc}
		\toprule
		&  scene  & objs & occlusion & movement & depth  & real    \\
		\midrule
		YCBV~\cite{xiangPoseCNN2018}   & static & multiple & heavy  & slow  & yes & yes \\
		BCOT~\cite{li2022bcot}   & dynamic & single & slight  & fast  & no & yes \\
		RBOT~\cite{tjadenRegionBasedGaussNewtonApproach2019}   & dynamic & single & slight  & fast  & no  & no \\
		\bottomrule
	\end{tabular}
	\caption{Properties of different datasets.}
	\label{tab:datasets}
\end{table}

\subsection{Datasets}

We use both YCBV and BCOT as our base datasets. They are both real-captured, and thus enable learning-based methods to be tested. In addition, their properties in scenes are highly complementary. As shown in \cref{tab:datasets}, YCBV is multi-object with realistic heavy occlusion, but it contains only static scenes with slow camera movements. On the contrary, BCOT is mainly for dynamic objects with fast moving and rotation, but contains only single-object scenes with slight occlusion. 
In fact, the properties of BCOT is quite similar as that of RBOT, which is semi-synthesized and may be biased for learning-based methods. Therefore, although RBOT is commonly used in previous works, in our benchmark we use BCOT instead.

BCOT provides 404 sequences of 21 different scenes; however, it does not split train and test sets. We thus select 11 scenes as the test set, with remaining 10 scenes as the train set. Although the train set is not as large as that of YCBV, this should not be a problem for 6DoF tracking since the 3D model has provide enough information.


\subsection{Subsequences}
\label{sec:subseq}

As discussed in \cref{sec:scoringm}, each sequence of the base datasets is split into multiple shorter subsequences in order for better AUC scoring. Each subsequence can be configured as $\{\xi^0,\mathcal{F}\}$, with $\xi^0$ the initial poses of tracking objects, and $\mathcal{F}$ the set of frames. The subsequence length $|\mathcal{F}|$ is randomly taken as 25, 50, 100, or 200. Shorter and longer subsequences are better for measuring  precision and robustness, respectively. The total number of frames for each length are made almost equal. Both forward and backward tracking are allowed for $\mathcal{F}$ in order to fully make use of the available frames.

Large inter-frame displacements~\cite{eccv2022} are generated by setting different frame steps. For YCBV and BCOT, the frame steps are randomly distributed in $[5, 15]$ and $[1, 4]$, respectively. The steps of YCBV are much larger because the original displacements are very small.

To test the case of unmodeled occlusion~\cite{tjadenRegionBasedGaussNewtonApproach2019}, for each subsequence of YCBV, we randomly remove up to one occluding object from the tracking list. Compared with the modeled occlusion, which can be inferred from the tracking objects~\cite{tjadenRegionBasedGaussNewtonApproach2019,huangOcclusionAwareEdge2020}, unmodeled occlusion is more difficult to be handled. The object \emph{036\_wood\_block} is removed from all sub-sequences due to the large reconstruction error (see the supplementary material).

We finally generate 401 and 1466 subsequences for YCBV and BCOT, respectively. The configurations of all susequences, together with required dataset informations, are stored as standard \emph{.json} files, so that successive works can easily reproduce the evaluations.

\subsection{Scoring and Ranking}
\label{sec:ranking}

As mentioned in section \ref{sec:errorm}, both ADD and PRJ errors need to be considered. For symmetric objects, ADD-S is used instead of ADD. This is essentially the ADD(-S) error~\cite{pengPVNetPixelWiseVoting2019,su2022zebrapose}. Only the objects with both shape and texture indistinguishable for some views are taken as symmetric (only the \emph{024\_bowl} of YCBV), in order to encourage methods that can better fuse all cues. The upper error bounds of ADD and PRJ for the AUC scores are set as $10cm$ and $10px$, respectively.

For the ranking, we finally compute a single score for each method by averaging the ADD and RRJ scores. In the following we refer this score as ADD-PRJ-AUC, which is taken as the overall accuracy of tracking methods.

  \begin{figure}
 	\centering
 	\small
 	\setlength\tabcolsep{0.5pt}
 	\begin{tabular}{ccc}
 		\includegraphics[width=0.33\linewidth]{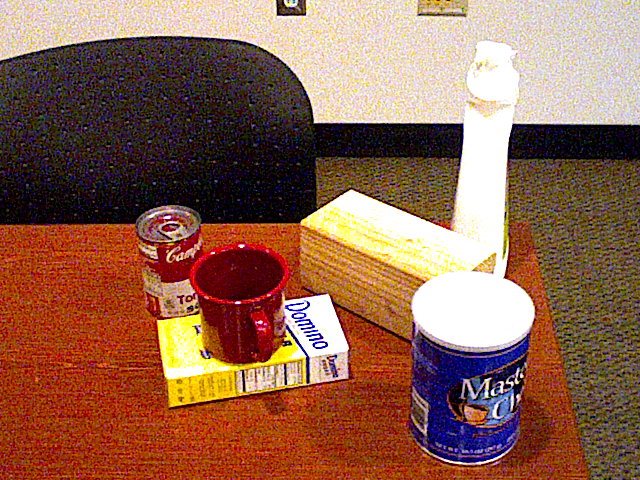} &
 		\includegraphics[width=0.33\linewidth]{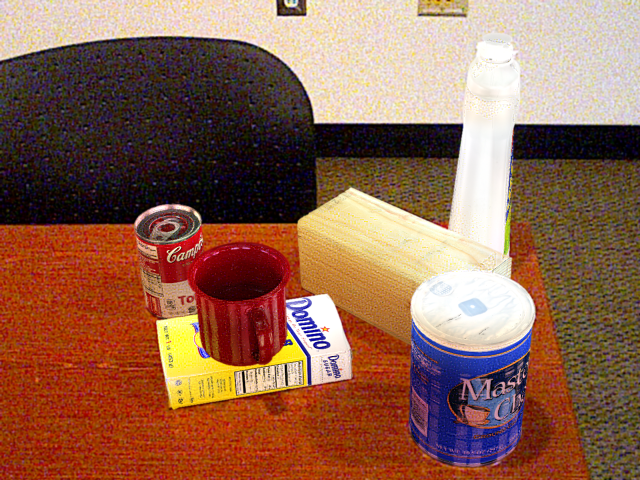}	&
 		\includegraphics[width=0.33\linewidth]{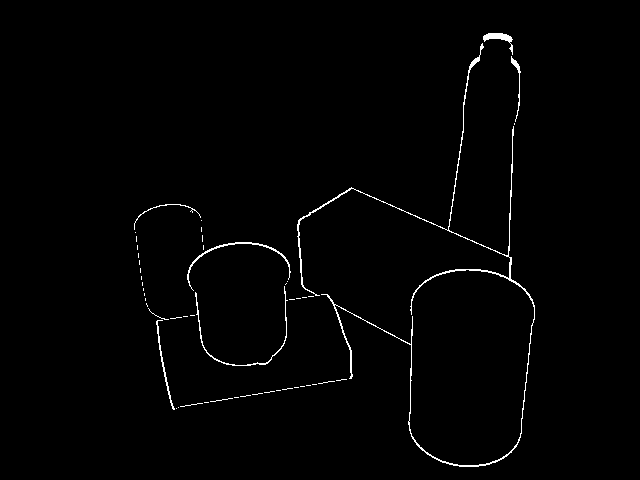} \\
 		(a) YCBV & (b) YCBV-SS & (c) dilate mask
 	\end{tabular}
 	\caption{The semi-synthesized YCBV variant (YCBV-SS).}
 	\label{fig:ycbvss}
 \end{figure}

 \begin{table}
 	\setlength\tabcolsep{2pt}
 	\small
 	\centering
 	\begin{tabular}{lccccccc}
 		\toprule
 		&  tx\text{\scriptsize(\emph{mm})}  & ty\text{\scriptsize(\emph{mm})} & tz\text{\scriptsize(\emph{mm})} & R\text{\scriptsize($^\circ$)}  & PRJ\text{\scriptsize(\emph{px})}  \\
 		\midrule
 		002\_master\_chef\_can            &0.28   &0.17   &1.11   &0.31   &0.50\\
 		003\_cracker\_box            &0.28   &0.20   &0.66   &0.07   &0.40\\
 		004\_sugar\_box            &0.19   &0.23   &0.81   &0.22   &0.38\\
 		005\_tomato\_soup\_can            &0.30   &0.21   &1.01   &0.15   &0.39\\
 		006\_mustard\_bottle            &0.18   &0.18   &1.05   &0.25   &0.41\\
 		007\_tuna\_fish\_can            &0.18   &0.22   &0.77   &0.20   &0.36\\
 		008\_pudding\_box            &0.25   &0.40   &0.58   &0.27   &0.54\\
 		009\_gelatin\_box            &0.18   &0.19   &0.64   &0.33   &0.28\\
 		010\_potted\_meat\_can            &0.27   &0.20   &0.62   &0.17   &0.40\\
 		011\_banana$^*$            &0.39   &0.31   &0.88   &0.40   &0.71\\
 		019\_pitcher\_base$^*$            &0.22   &0.17   &0.70   &0.21   &0.45\\
 		021\_bleach\_cleanser            &0.25   &0.26   &1.00   &0.26   &0.47\\
 		024\_bowl$^*$            &0.39   &0.19   &0.50   &0.26   &0.67\\
 		025\_mug$^*$            &0.42   &0.36   &1.25   &0.34   &0.62\\
 		035\_power\_drill            &0.27   &0.19   &0.86   &0.18   &0.41\\
 		037\_scissors$^*$            &0.25   &0.20   &0.97   &0.25   &0.45\\
 		040\_large\_marker            &0.23   &0.25   &0.72   &0.34   &0.42\\
 		051\_large\_clamp$^*$            &0.40   &0.12   &0.85   &0.22   &0.51\\
 		052\_extra\_large\_clamp$^*$           &0.35   &0.30   &0.75   &0.21   &0.59\\
 		061\_foam\_brick$^*$            &0.49   &0.18   &0.58   &0.44   &0.71\\
 		all            &0.28   &0.22   &0.85   &0.23   &0.46\\
 		\midrule
 		binocular-$90^\circ$~\cite{li2022bcot} & 0.28 & 0.23 & 0.28 & 0.76 & - \\
 		binocular-$45^\circ$~\cite{li2022bcot} & 0.57 & 0.28 & 1.35 & 0.94 & - \\
 		YCBV original    & 3.03 & 2.15 & 21.4 & 3.27 & 4.33 \\
 		\bottomrule
 	\end{tabular}
 	\caption{The evaluations of bundle pose refinement. The textureless objects are marked with $*$.}
 	\label{tab:brp_evals}
 \end{table}
 
 \begin{table}
 	\setlength\tabcolsep{4pt}
 	\small
 	\centering
 	\begin{tabular}{lccccccc}
 		\toprule
 		&  tx  & ty & tz & R  & PRJ  \\
 		\midrule
 		w/o $f^{edge}$   & 0.29 & 0.25 & 1.40 & 0.29 & 0.51 \\
 		w/o $f^{point}$ & 0.34 & 0.32 & 1.96 & 0.36 & 0.60 \\
 		w/o $f^t$     & \textbf{0.28} & 0.23 & 1.16 & 0.26 & 0.48   \\
 		using all        & \textbf{0.28}  & \textbf{0.22}  & \textbf{0.85} & \textbf{0.23}  & \textbf{0.46}\\
 		\bottomrule
 	\end{tabular}
 	\caption{The ablation studies to our refinement method.}
 	\label{tab:ablations}
 \end{table}

\section{Experiments}

In experiments we first evaluate the proposed bundle pose refinement method, and then present the benchmark results with some analysis.

\subsection{Evaluation of Bundle Pose Refinement}

\paragraph{Dataset}
In order for this evaluation, a dataset with accurate groundtruth is required. We thus generate a semi-synthesized YCBV variant as shown in \cref{fig:ycbvss}, which replaces the original object regions by rendered objects w.r.t the refined poses, with 3D models slightly enlarged by 2\%-5\% (the dilate mask in \cref{fig:ycbvss}), so that the original regions can be completely covered. Color transformations and noise are also applied for more realistic appearance. Please see the supplementary material for details.


We call the above YCBV variant as YCBV-SS, which also can be used by successive works when accurate groundtruth is required. 

\newcommand{\RGB}{\text{\scriptsize RGB}}
\newcommand{\RGBD}{\text{\scriptsize RGBD}}
\newcommand{\larger}[1]{\text{\footnotesize #1}}
\newcommand{\first}[1]{\textbf{#1}}

\renewcommand\arraystretch{1.0}
\begin{table*}[h]
	\scriptsize
	\centering
		\setlength\tabcolsep{2.2pt}
			\begin{tabular}{ccccccccccccccccccccccccc}
				\toprule
			\larger{Method}	& Input & \text{002} & \text{003} & \text{004} & \text{005} & \text{006} & \text{007} & \text{008} & \text{009} & \text{010} & \text{011} & \text{019} & \text{021} & \text{024} & \text{025} & \text{035} & \text{037} & \text{040} & \text{051} & \text{052} & \text{061} & \larger{Avg.$^\uparrow$} & ADD & PRJ  \\
				\midrule
\larger{RBOT~\cite{tjadenRegionBasedGaussNewtonApproach2019} }& \RGB &22.9  &28.1  &30.4  &15.6  &18.6  &22.3  &8.9  &32.0  &20.9  &12.7  &18.4  &13.9  &38.7  &20.9  &18.6  &9.6  &8.6  &7.3  &12.4  &25.4  
& \larger{19.3}  & 33.0$^{[12]}$ & 5.60$^{[12]}$ \\
\larger{PoseRBPF$^\dagger$~\cite{dengPoseRBPF2021}} & \RGB &32.8  &33.5  &43.3  &36.8  &47.3  &27.0  &29.3  &53.2  &24.5  &20.2  &35.2  &19.2  &47.6  &46.9  &38.7  &23.3  &40.9  &32.8  &29.4  &31.7  
& \larger{34.7} & 50.2$^{[11]}$ & 19.1$^{[10]}$ \\
\larger{PoseRBPF$^\dagger$~\cite{dengPoseRBPF2021}} & \RGBD &44.2  &43.3  &48.4  &43.8  &46.4  &33.4  &35.0  &52.8  &33.7  &46.2  &42.1  &39.5  &42.7  &51.6  &48.7  &41.6  &46.3  &41.8  &35.8  &46.8  
& \larger{43.2} & 70.1$^{[4]}$ & 16.3$^{[11]}$ \\
\larger{DeepIM$^\dagger$~\cite{lideepim2018}} & \RGB &54.0  &44.1  &60.4  &45.0  &42.8  &46.0  &22.2  &58.2  &41.3  &38.9  &46.5  &46.6  &39.1  &57.2  &56.6  &37.9  &32.1  &38.5  &28.2  &38.9  
&\larger{43.7}  & 58.4$^{[10]}$ & 29.1$^{[8]}$\\
\larger{ICG~\cite{stoiber2022iterative}} & \RGB  &39.1  &37.4  &59.3  &34.3  &49.2  &44.6  &39.6  &58.4  &48.3  &46.1  &53.9  &45.2  &47.0  &55.4  &42.9  &41.6  &28.0  &52.1  &35.0  &48.2  
&\larger{45.3}  & 58.5$^{[9]}$ & 32.0$^{[6]}$\\
\larger{DeepIM$^\dagger$~\cite{lideepim2018}} & \RGBD &\first{57.1}  &42.5  &60.6  &45.9  &44.4  &49.7  &29.2  &55.2  &37.4  &47.8  &53.1  &46.6  &38.6  &60.7  &55.7  &38.9  &42.0  &46.5  &32.7  &42.7  
& \larger{46.4} & 64.2$^{[7]}$ & 28.6$^{[9]}$ \\
\larger{TrackNet$^\dagger$~\cite{wen2020se}} & \RGBD &53.8  &41.3  &66.5  &57.1  &62.4  &34.0  &40.4  &46.7  &35.9  &51.0  &69.8  &55.4  &46.5  &42.5  &56.7  &43.8  &48.2  &40.9  &39.3  &14.9  
& \larger{47.4} & 64.3$^{[6]}$ & 30.5$^{[7]}$ \\
\larger{LDT~\cite{eccv2022}} & \RGB  &30.4  &41.1  &76.8  &41.6  &\first{88.9}  &46.4  &12.9  &88.7  &39.1  &46.4  &85.0  &61.9  &51.9  &77.6  &46.6  &15.0  &22.3  &48.6  &37.1  &71.9  
& \larger{51.5}  & 62.9$^{[8]}$ & 40.1$^{[5]}$ \\
\larger{RBGT~\cite{stoiber2020sparse}} & \RGB &45.8  &44.4  &79.9  &43.1  &81.4  &46.5  &34.0  &54.8  &31.3  &66.5  &83.5  &45.4  &55.6  &83.6  &67.5  &47.2  &25.1  &71.8  &27.0  &66.0  
& \larger{55.0} & 67.3$^{[5]}$ & 42.7$^{[4]}$ \\
\larger{SRT3D~\cite{stoiber2021srt3d}} & \RGB &46.5  &50.2  &83.6  &46.6  &85.5  &47.4  &51.1  &83.1  &43.2  &76.5  &85.2  &52.0  &54.4  &77.8  &76.3  &55.3  &23.7  &74.0  &41.9  &69.4  
& \larger{61.2}  & 73.3$^{[3]}$ & 49.1$^{[3]}$ \\
\larger{ICG~\cite{stoiber2022iterative}} & \RGBD &48.9  &\first{65.0}  &77.2  &54.3  &76.2  &54.8  &35.1  &78.5  &62.7  &79.8  &75.7  &62.5  &57.5  &72.7  &78.7  &55.6  &\first{55.5}  &68.2  &50.1  &77.8  
& \larger{64.3} & 76.1$^{[2]}$ & 52.6$^{[2]}$ \\
\larger{ICG+SRT3D} & \RGBD &55.8  &63.7  &\first{86.0}  &\first{56.0}  &86.1  &\first{61.3}  &\first{42.7}  & \first{89.4}  &\first{64.6}  &\first{82.2}  &\first{88.3}  &\first{68.3}  &\first{61.1}  &\first{83.6}  &\first{89.6}  &\first{61.9}  &53.5  &\first{78.1}  &\first{53.3}  &\first{81.5}  &\larger{70.3} & 83.8$^{[1]}$ & 56.9$^{[1]}$ \\

\bottomrule
\end{tabular}
\caption{The benchmark results on YCBV with our refined annotations. ADD-PRJ-AUC is used as the accuracy. The last two columns also show the ADD and PRJ scores, with ranks in the brackets. The methods marked with $\dagger$ are learning-based.}
\label{tab:ycbv_benchmark}
\end{table*}

\begin{table*}[ht]
	\scriptsize
	\centering
	\setlength\tabcolsep{2.5pt}
			\begin{tabular}{ccccccccccccccccccccccccc}
				\toprule
				\larger{Method} & Input & \rotatebox{60}{\text{touch}} & \rotatebox{60}{\text{ape}} & \rotatebox{60}{\text{gps}} & \rotatebox{60}{\text{bracket}} & \rotatebox{60}{\text{cat}} & \rotatebox{60}{\text{pool}} & \rotatebox{60}{\text{driller}} & \rotatebox{60}{\text{light}} & \rotatebox{60}{\text{jack}} & \rotatebox{60}{\text{clamp}} & \rotatebox{60}{\text{lego}} & \rotatebox{60}{\text{clip}} & \rotatebox{60}{\text{arm}} & \rotatebox{60}{\text{squirrel}} & \rotatebox{60}{\text{tube}} & \rotatebox{60}{\text{stitch}} & \rotatebox{60}{\text{teapot}} & \rotatebox{60}{\text{tube}} & \rotatebox{60}{\text{queen}} & \rotatebox{60}{\text{shelf}} & \larger{Avg.$^\uparrow$} & ADD & PRJ \\
				\midrule
\larger{RBOT~\cite{tjadenRegionBasedGaussNewtonApproach2019}}  & \RGB 
				  &7.1  &6.7  &9.3  &9.8  &12.5  &8.4  &17.0  &4.3  &6.1  &11.3  &7.4  &5.9  &3.6  &8.7  &10.2  &7.5  &10.0  &9.0  &10.2  &7.6  &\larger{8.6} & 11.3$^{[5]}$ & 5.9$^{[5]}$  \\
			\larger{ICG~\cite{stoiber2022iterative}}  &\RGB &17.6  &28.2  &22.0  &24.8  &36.3  &23.8  &41.3  &18.0  &15.8  &33.4  &34.0  &20.8  &20.7  &31.4  &27.6  &18.3  &26.2  &26.1  &38.3  &28.0  &\larger{26.6} & 30.4$^{[4]}$ & 22.9$^{[4]}$   \\
				 \larger{RBGT~\cite{stoiber2020sparse}} &\RGB  &21.1  &37.4  &35.4  &30.5  &42.6  &38.1  &45.7  &25.3  &19.3  &42.0  &28.0  &25.2  &21.6  &45.2  &30.6  &25.8  &35.8  &39.8  &41.0  &27.2  &\larger{32.9} & 35.9$^{[3]}$ & 29.8$^{[3]}$ \\
				 \larger{SRT3D~\cite{stoiber2021srt3d}}  &\RGB &29.2  &48.1  &44.4  &42.9  &56.8  &46.9  &55.5  &34.5  &29.4  &48.3  &39.0  &34.8  &26.6  &50.9  &36.2  &33.2  &42.8  &48.3  &49.1  &35.2  &\larger{41.6} & 44.8$^{[2]}$ & 38.5$^{[2]}$ \\
				 \larger{LDT~\cite{eccv2022}}  &\RGB &\textbf{50.0}  &\textbf{80.3}  &\textbf{60.9}  &\textbf{76.2}  &\textbf{76.3}  &\textbf{68.2}  &\textbf{75.0}  &\textbf{60.3}  &\textbf{46.4}  &\textbf{76.8}  &\textbf{67.7}  &\textbf{48.8}  &\textbf{60.0}  &\textbf{70.0}  &\textbf{76.9}  &\textbf{52.4}  &\textbf{74.7}  &\textbf{78.2}  &\textbf{66.2}  &\textbf{59.3}  &\larger{66.2} & 68.7$^{[1]}$ & 63.8$^{[1]}$  \\
				\bottomrule
			\end{tabular}
\caption{The benchmark results on BCOT for representative non-learning RGB-based methods. }
\label{tab:bcot_benchmark}
\end{table*}

\begin{table}
	\setlength\tabcolsep{4pt}
	\small
	\centering
	\begin{tabular}{lccccccc}
		\toprule
		\multicolumn{2}{c}{Method}  &  \multicolumn{3}{c}{YCBV Original} & \multicolumn{3}{c}{Our Refined}\\
		& && {\scriptsize ADD} & {\scriptsize PRJ} && {\scriptsize ADD} & {\scriptsize PRJ} \\
		\midrule
		PoseRBPF~\cite{dengPoseRBPF2021} & \RGBD && 81.9 & 17.6 && 70.1 & 16.3 \\
		TrackNet~\cite{wen2020se} & \RGBD   &&  75.5 & 28.4 && 64.3 & 30.5 \\
		DeepIM~\cite{lideepim2018} & \RGBD   & & 76.8 & 29.9 && 64.2 & 28.6 \\
		ICG~\cite{stoiber2022iterative} & \RGBD   && 91.1 & 45.2 && 76.1 & 52.6 \\
		
		\midrule
		SRT3D~\cite{stoiber2021srt3d} & \RGB && 65.3 & 33.7 && 73.3 & 49.1 \\
		RBGT~\cite{stoiber2020sparse} & \RGB && 61.3 & 30.6 && 67.3 & 42.7 \\
		LDT~\cite{eccv2022}  & \RGB && 54.8 & 26.2 && 62.9 & 40.1 \\
		ICG+SRT3D  & \RGBD  && 79.7 & 39.6  && 83.8 & 56.9 \\ 
		\bottomrule
	\end{tabular}
	\caption{Results with different annotations for YCBV.}
	\label{tab:compare_annot}
\end{table}

\begin{figure*}
	\centering
	\setlength\tabcolsep{0pt}
	\begin{tabular}{cccc}
		\includegraphics[width=0.2475\linewidth]{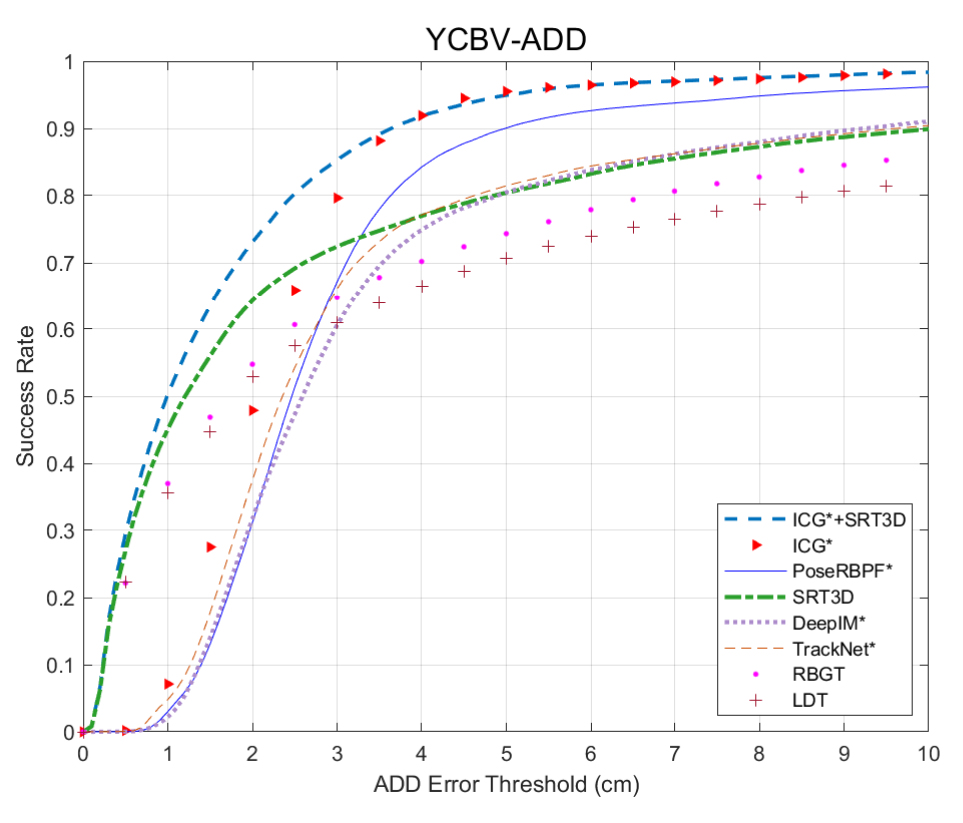} &
		\includegraphics[width=0.2475\linewidth]{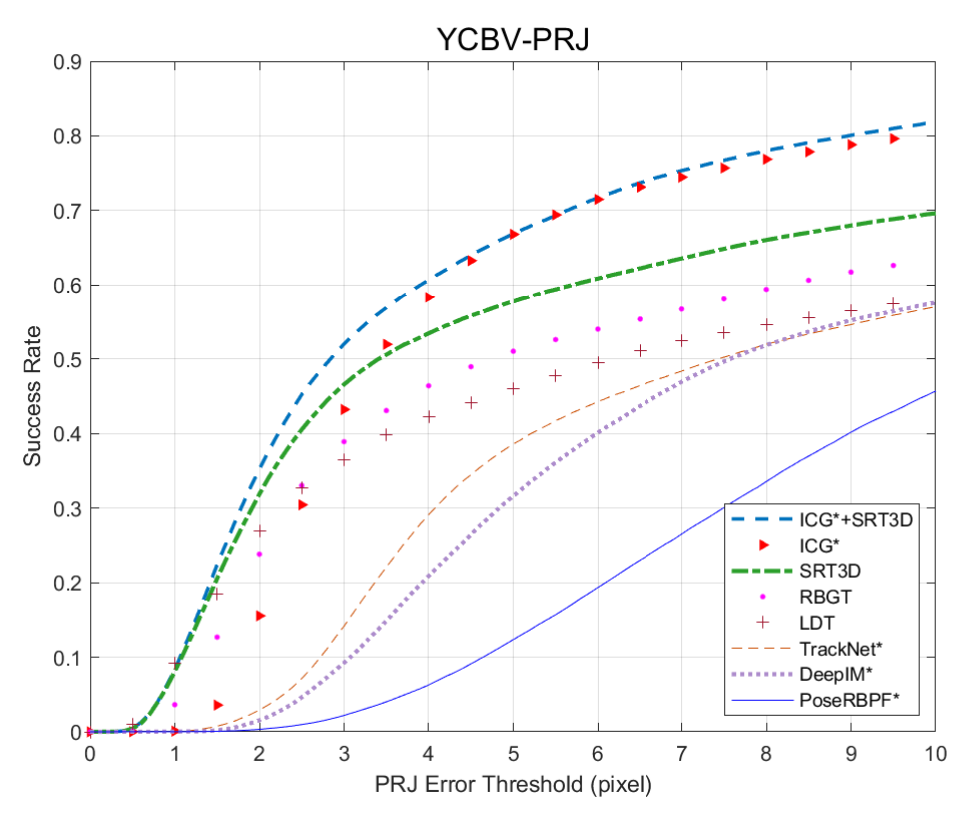}	&	
		\includegraphics[width=0.2475\linewidth]{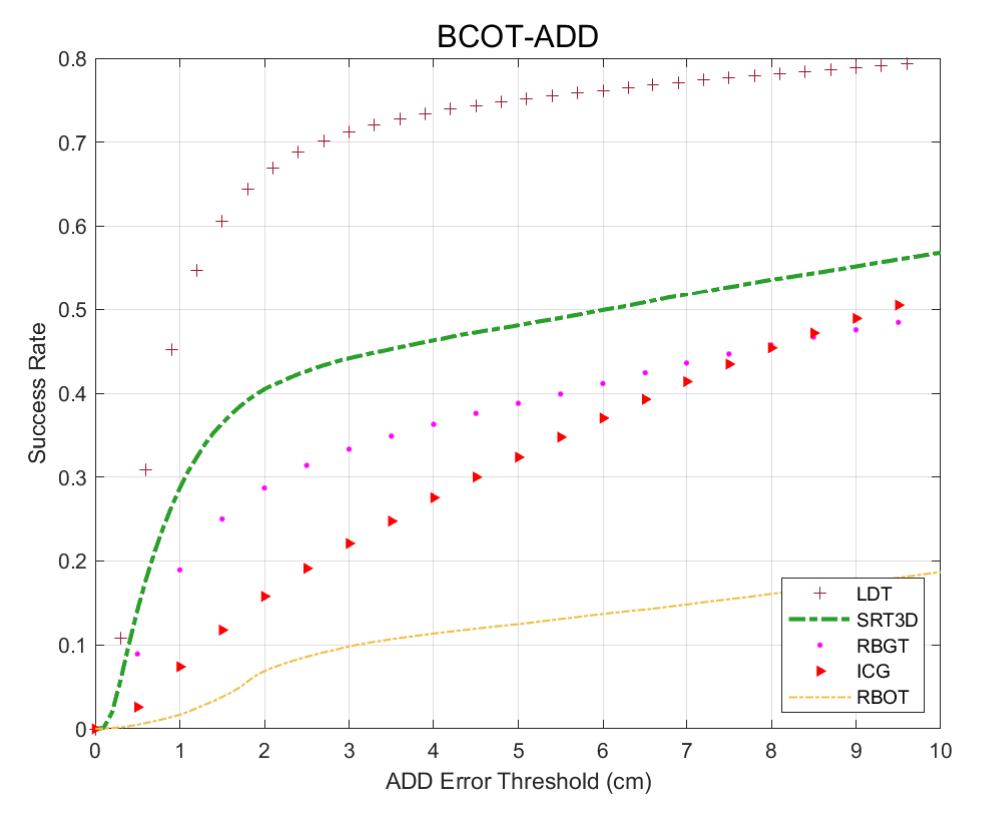} &
		\includegraphics[width=0.2475\linewidth]{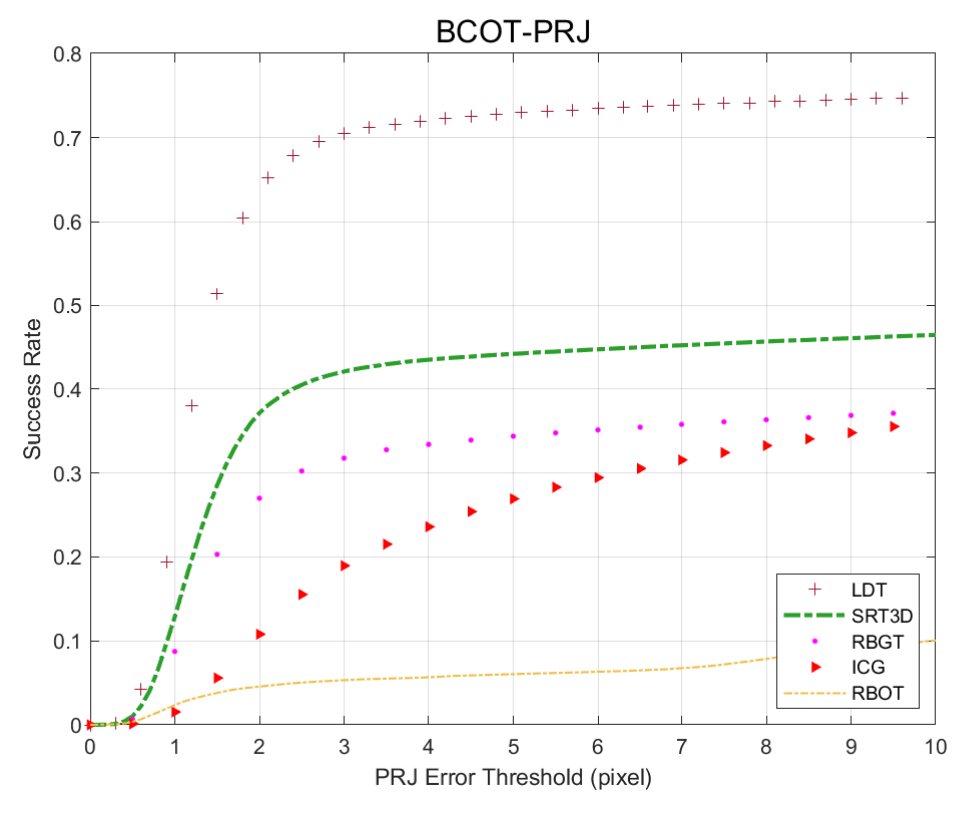}
	\end{tabular}
	\caption{The accuracy curves. For YCBV only top-performing methods are included, and the methods with * are RGBD-based.}
	\label{fig:curve}
\end{figure*}

\paragraph{Results}
\cref{tab:brp_evals} shows the accuracy of our refinement method on YCBV-SS. The initial poses are generated by adding a random deviation to the original YCBV annotations, in order to test similar case as to refine YCBV. The average ADD and PRJ errors of the initial poses are $22.1mm$ and $5.2px$, respectively. As can be seen, after the refinement we obtain average translation error less than $1mm$ and PRJ error less than $0.5px$. The accuracy of individual objects vary a little, but there is no abnormal case, textured and textureless objects also take no obvious difference in accuracy, showing the robustness and reliability of our method to different situations. 

\cref{tab:brp_evals} also lists the results of the binocular method in \cite{li2022bcot}. Note that the results are from their own synthetic experiments with different dataset, so it is just for a reference. As is shown, our method can achieve comparable accuracy with the best case (binocular-$90^\circ$) of \cite{li2022bcot}. To be more critical, our method results in smaller rotation error but larger translation error. This should be mainly because that the estimation of accurate translation is largely dependent on the camera poses, which are accurate in the synthetic experiments of \cite{li2022bcot}, while in our method the camera poses are automatically estimated. 

The last row of \cref{tab:brp_evals} is the difference of the YCBV orignal annotations with our refined poses. Since the error of our method is very small in comparison with this difference, we can take this difference as the error of the original YCBV annotations. The translation error is more than $2cm$, which would largely influence the resulting scores (see below). Actually, as mentioned in \cite{li2022bcot}, the random depth error of Kinect DK may be up to $17mm$, so it should not be surprising for the large error of the original annotations.


\cref{tab:ablations} shows the ablation studies to our method. All of three loss functions contribute positively to the result accuracy, and removing either of them could prevent our method to achieve sub-millimeter precision.

\subsection{The Benchmark Results}
\label{sec:benchmark_results}

\paragraph{RGB \& RGBD} \Cref{tab:ycbv_benchmark} and \ref{tab:bcot_benchmark} are our benchmark results for representative 6DoF pose tracking methods. For YCBV, our refined annotations are used as the groundtruth for both RGB and RGBD methods. Therefore, our reported accuracy for previous RGBD methods should be underestimated since they are tuned according to the original depth-based annotations. However, we argue that considering the depth error and color-depth misalignments, it should be better to solve the pose w.r.t the color image, while using depth mainly for improving the robustness. For a quick test, we refine the results of ICG with SRT3D, and obtain the row ICG+SRT3D in \cref{tab:ycbv_benchmark}, which can be taken as the start-of-the-art accuracy of RGBD methods for the benchmark.

\vspace{3pt}
\noindent\textbf{Original \& Refined Annotations} \cref{tab:compare_annot} compares the accuracy computed with YCBV original annotations and our refined ones. For RGBD methods, the ADD scores with our refined annotations are significantly lower, while the PRJ scores are still comparable, and even better for ICG and TrackNet, which are shown to be better aligned with RGB images. On the other hand, for RGB-based methods, both ADD and PRJ scores are significantly better with our annotations. In particular, with the original annotation ICG+SRT3D performs significant worse than ICG, but actually this is not the truth according our visual analysis (see the supplementary material).


\vspace{3pt}
\noindent\textbf{ADD \& PRJ} The last two columns of \cref{tab:ycbv_benchmark} and \ref{tab:bcot_benchmark} show the results of ADD and PRJ scores. We can find that the corresponding ranks are somewhat different, showing the difference of trackers in robustness and precision. Typically, for PoseRBPF with RGBD, the ranks differ a lot (4 \emph{v.s.} 11). According to the accuracy curves shown in \cref{fig:curve}, this is because PoseRBPF has high robustness but low precision. It is the 2nd best for ADD error threshold $t>3cm$, but is almost the worst for $t<3cm$. As analyzed in \cref{sec:errorm}, using only ADD  in this case may produce biased scores.

\vspace{3pt}
\noindent\textbf{YCBV \& BCOT} The results of LDT~\cite{eccv2022} differs a lot for the two datasets. LDT does not perform well on YCBV, but significantly outperform other methods on BCOT. This is mainly due to occlusion. The method of LDT does not consider occlusion, so it is sensitive to heavy occlusions. For example, several objects (\emph{008\_pudding\_box}, \emph{037\_scissors}, etc.) are occluded more than half all the time, for which LDT achieves very poor accuracy. This problem is not exposed on BCOT and RBOT, whose occlusion case presents only slight occlusions. On the other hand, YCBV contains only static scenes with small frame displacements, so the non-local search of LDT could not take effect, and it is beneficial for methods with only local optimization, which however, would be problematic for the fast object movement of BCOT.

\vspace{3pt}
\noindent\textbf{Learning \& Non-learning} Both learning and non-learning methods are included in \cref{tab:ycbv_benchmark}. Only the methods with public code and trained model are tested, so there is no method available for BCOT. Currently, learning-based methods do not perform as well as optimization-based methods, as has been evidenced in \cite{stoiber2022iterative}. This should be because 6DoF tracking requires only local search in the pose space, which can be well constrained by the initial pose and 3D model information, so it is less dependent on priors learned from the training data. However, we believe that for more challenge case with the problem under constrained, it should be more beneficial to take the learning-based methods.

\section{Conclusions}

In this paper we contribute datasets and methods for a unified comprehensive evaluation of 6DoF object pose tracking. The proposed bundle pose refinement method can achieve sub-pixel and sub-millimeter errors, based on which we re-annotate the YCBV dataset. We experimentally estimate the errors of the original annotations, which are shown to be unreliable for high-precision tracking. Although in this paper we focus on only 6DoF tracking, similar problems also exist for 6DoF pose refinement~\cite{lideepim2018,iwase2021repose,lipson2022coupled,xu2022rnnpose} and 6DoF pose estimation~\cite{xiangPoseCNN2018,pengPVNetPixelWiseVoting2019,su2022zebrapose}, which also commonly use YCBV for their evaluations, and thus can benefit from our work.


Based on YCBV and BCOT datasets, we introduce a new unfied benchmark with standard protocol and improved scoring methods. Some unknown properties of previous methods are discovered and analyzed. In particular, we show the large influence of depth error, and demonstrate a better way with depth mainly for improving robustness. The limitation of the widely-used ADD error metric is discovered and evidenced. We also find that there is no common winner in YCBV and BCOT datasets, which clearly show the limitations of related tracking methods. It also reveals some immediate chances to get improvements, which can be considered by future 6DoF object tracking works.



{\small
\bibliographystyle{ieee_fullname}
\bibliography{egbib}
}

\end{document}